\documentclass{article}
\usepackage{microtype}
\usepackage{graphicx}
\usepackage{booktabs}
\usepackage{hyperref}
\usepackage[colorinlistoftodos]{todonotes}
\usepackage{comment}
\usepackage{soul}
\usepackage{multirow}
\usepackage{amssymb}
\usepackage{xspace}
\usepackage{subcaption}

\newcommand{\lucia}[1]{\todo[color=red!40]{#1}}

\usepackage[accepted]{icml2019}
\newcommand{\org}{\texttt{ORG}\xspace}
\newcommand{\act}{\texttt{ACT}\xspace}
\newcommand{\all}{\texttt{ALL}\xspace}
\newcommand{\bleu}{\texttt{BLEU}\xspace}
\newcommand{\aic}{\texttt{AIC-videosum}\xspace}
\newcommand{\conv}{\texttt{AIF-conv4}\xspace}
\newcommand{\videosum}{\texttt{AIF-videosum}\xspace}
\newcommand{\emb}{\texttt{AIF-emb}\xspace}

\newcommand{\verbmask}{\texttt{\textcolor{red}{\textbf{V}}} }

\graphicspath{{figures/}}

\icmltitlerunning{Predicting Actions to Help Predict Translations}

\begin{document}

\twocolumn[
\icmltitle{Predicting Actions to Help Predict Translations}

\icmlsetsymbol{equal}{*}

\begin{icmlauthorlist}
\icmlauthor{Zixiu Wu}{ic}
\icmlauthor{Julia Ive}{shef}
\icmlauthor{Josiah Wang}{ic}
\icmlauthor{Pranava Madhyastha}{ic}
\icmlauthor{Lucia Specia}{ic}
\end{icmlauthorlist}

\icmlaffiliation{ic}{Department of Computing, Imperial College London, United Kingdom}
\icmlaffiliation{shef}{DCS, Sheffield University, United Kingdom}

\icmlcorrespondingauthor{Lucia Specia}{l.specia@imperial.ac.uk}

\icmlkeywords{Multimodality, Multimodal Machine Translation, Computer Vision, Action Recognition, Vision and Language}

\vskip 0.3in
]

\printAffiliationsAndNotice{}

\begin{abstract}
We address the task of text translation on the How2 dataset using a state of the art transformer-based multimodal approach. 
The question we ask ourselves is whether visual features can support the translation process, in particular, 
given that this is a dataset extracted from videos, we focus on the translation of actions, which we believe are poorly captured in current static image-text datasets currently used for multimodal translation.
For that purpose, we extract different types of action features from the videos and carefully investigate how helpful this visual information is by testing whether it can increase translation quality when used 
in conjunction with (i) the original text and (ii) the original text where action-related words (or all verbs) are masked out. The latter is a simulation that helps us assess the utility of the image in cases where the text does not provide enough context about the action, or in the presence of noise in the input text.

\end{abstract}

\section{Introduction}
\label{sec:introduction}

Multimodal machine translation (MMT) \cite{SpeciaEtAl:2016} is one of the main applications motivating the creation of the How2 dataset \cite{how2-dataset}. 
The goal was to move away from existing datasets -- namely Multi30K \cite{elliott-etall_VL:2016} -- with static images and their corresponding simple and short descriptive captions. 
In the Multi30K dataset, existing work has shown that images can be beneficial, especially in the presence of noisy or incomplete input \cite{naacl19-probing,ive2019distil}.

The language in the How2 dataset is not necessarily descriptive and sentences are longer, less repetitive and structurally more complex. 
While intuitively this should make the translation task harder and under such conditions one could expect that other modalities could be helpful, the general translation quality obtained by text-only neural machine translation models trained on this dataset is relatively high, as reported in \cite{how2-dataset}. 
Additionally, there is not a very close equivalence between the visual and textual modality. For example, many videos are focused on the speaker. Therefore, making use of the additional modality becomes a much harder challenge.
As a consequence, previous experiments on MMT on this data thus far have not been able to benefit from images \cite{how2-dataset}.
In this paper we further examine the question of whether visual information can be helpful by (i) using a more advanced model architecture for multimodality, (ii) testing different types of visual features and and different ways of representating these features; and (iii) concentrating on the translation of words which we believe the temporal nature of videos could help with.

More specifically, in a similar way to \citet{naacl19-probing}, we probe the contribution of images by masking source words to simulate the case of noisy or highly ambiguous input. We focus on actions, which are generally represented by certain verbs,
as we believe this is the main additional information one can explore in videos, as compared to static images. We report experiments with a more advanced, transformer-based architecture for MMT than that exploited in \citet{how2-dataset}. Our results show that the visual features, especially those from a CNN fine-tuned for classifying videos into verb-related actions, can be beneficial, in particular for masking settings. 
Human evaluation of a subset of the data confirms the automatic evaluation results.

\section{Dataset and Masking Strategies}

We use the How2 \cite{how2-dataset} dataset for the experiments, keeping the standard splits:\footnote{\url{https://github.com/srvk/how2-dataset}} 184,949 training sentences, 2,022 validation sentences and 2,305 test sentences. 
Our text-only baseline uses the dataset as distributed. For the masking experiments, two strategies to replace words in the source language are defined:
\begin{itemize}
\item \textbf{Mask action verbs (\act):} All verbs which correspond to an action as defined in the action categorisations of the Moments in Time dataset~\cite{monfortmoments}
are replaced by a placeholder. The masked words (tokens) make up 2.75\%, 2.83\%, and 2.84\% of the training, validation, and test texts respectively. 
\item \textbf{Mask all verbs (\all):} All verbs in the sentence are replaced by a placeholder. The masked words (tokens) make up 20.6\%, 21.0\%, and 20.4\% of the training, validation, and test texts respectively. 
\end{itemize}

The masking is performed in all sentences containing (action) verbs in the source language. For that, the data is first POS-tagged and lemmatised using spaCy 2.0.\footnote{\url{http://spacy.io/} model \texttt{en\_core\_web\_lg}} In the case of action verbs, the resulting lemmatised tokens are matched against the 339 lemmatised action verbs from \citet{monfortmoments}.\footnote{We retain only the verb component for specialised actions such as \textit{playing+music} and \textit{adult+male+singing}} The target language remains the same for the purposes of both training and testing. We call the original unmasked sentences \org.

\begin{figure*}[th]
    \centering
    \small
    \begin{minipage}{0.25\textwidth}
        \includegraphics[height=2cm]{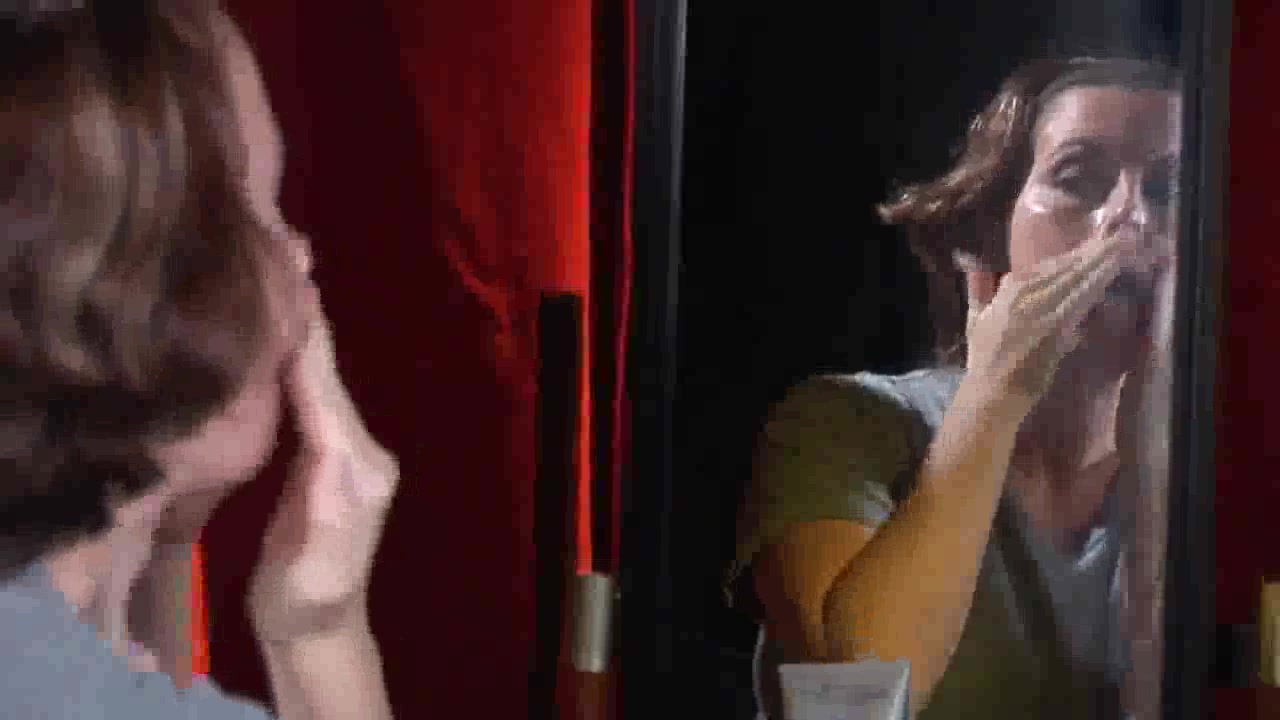}
    \end{minipage}
    \begin{minipage}{0.70\textwidth}
        $\blacksquare$ simply apply the cleanser or cream to your hands and apply it to the face and begin rubbing. \\
        $\blacklozenge$ simply apply the cleanser or cream to your hands and apply it to the face and begin \verbmask .\\
        $\blacktriangle$ simply \verbmask the cleanser or cream to your hands and \verbmask it to the face and \verbmask \verbmask .
    \end{minipage}
    
    \begin{minipage}{0.25\textwidth}
        \includegraphics[height=2cm]{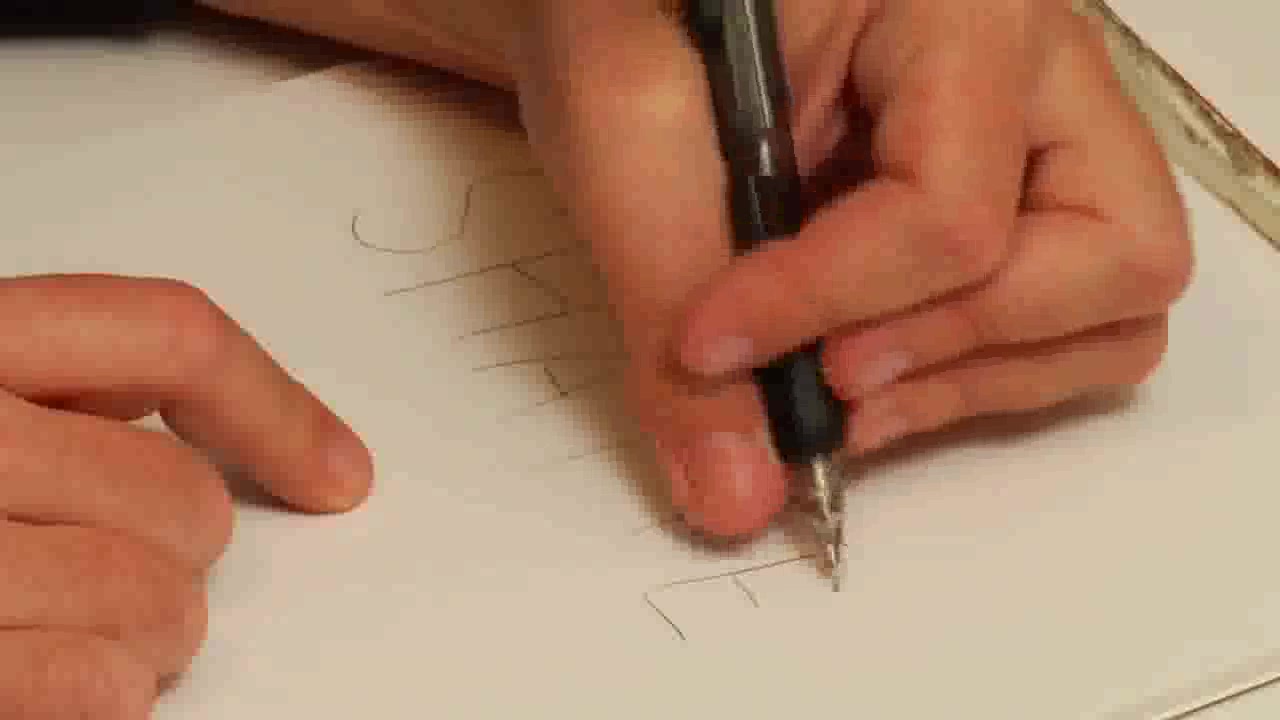}
    \end{minipage}
    \begin{minipage}{0.70\textwidth}
        $\blacksquare$ you can draw it really lightly , go back and erase it later .\\
        $\blacklozenge$ you can \verbmask it really lightly , go back and erase it later .\\
        $\blacktriangle$ you \verbmask \verbmask it really lightly , \verbmask back and \verbmask it later .
    \end{minipage}   
    
    \begin{minipage}{0.25\textwidth}
        \includegraphics[height=2.5cm]{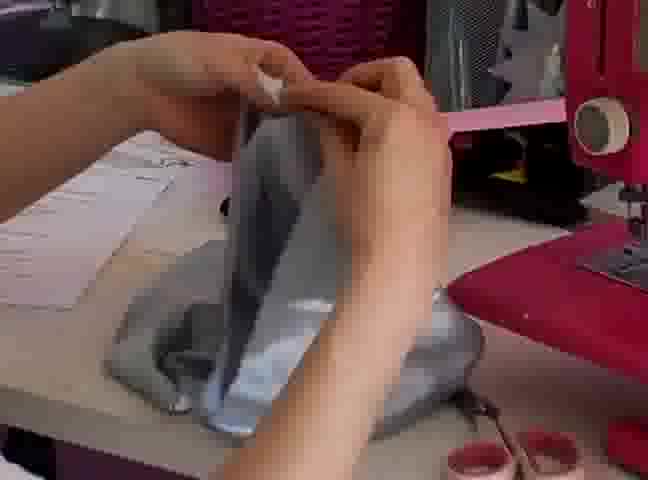}
    \end{minipage}
    \begin{minipage}{0.70\textwidth}
         $\blacksquare$ what we are going to be doing is folding the top over and making a little casing the ribbon will slip through .\\
         $\blacklozenge$ what we are going to be doing is \verbmask the top over and making a little casing the ribbon will \verbmask through .\\
         $\blacktriangle$ what we \verbmask \verbmask to \verbmask \verbmask \verbmask \verbmask the top over and \verbmask a little \verbmask the ribbon \verbmask \verbmask through .
    \end{minipage}

    \caption{Three example segments from the How2 training dataset with verbs masked. In each example, the first line ($\blacksquare$) shows the full text segment, the second line ($\blacklozenge$) shows the segment with verbs from \citet{monfortmoments} masked with \verbmask, the third line ($\blacktriangle$) shows the segment with all verbs masked with \verbmask.}
    \label{fig:egmasked}
\end{figure*}

Figure~\ref{fig:egmasked} shows some examples of segments from How2 with verbs masked using the two different strategies.

Byte Pair Encoding (BPE) \cite{sennrich2015neural} with 20,000 merge operations is applied on the target training text and each of the differently-masked source training texts separately, leading to 4 distinct vocabularies for \org, \all, \act, and the target language respectively. 

\section{Visual features} \label{visual features}

We experiment with three types of visual features:
\begin{itemize}
    \item \textbf{videosum:} the output of the last fully-connected layer of ResNeXt-101 \cite{xie2017aggregated} with 3D convolutional kernels trained to recognise 400 different actions \cite{hara2018can};
    \item \textbf{conv4:} the final convolutional layer of a 3D ResNet-50 CNN trained to classify the 339 action verbs from \citet{monfortmoments}; 
    \item \textbf{emb}: a word embedding matrix for the 339 action verbs, with the embedding of each verb weighted by the final softmax layer of the same CNN for \textbf{conv4}.
\end{itemize}

\textbf{videosum} is provided officially by the How2 Challenge.\footnote{\url{https://srvk.github.io/how2-challenge/}} Each How2 video segment is divided into 16-frame chunks as separate inputs to the network, according to \citet{how2-dataset}, and the average of the 2048-D feature maps for all the chunks is computed as the single-vector feature of the video segment.

We extract \textbf{conv4} and \textbf{emb} features using a 3D ResNet-50 CNN trained by \citet{monfortmoments}, which inflates a 2D ResNet-50 CNN pre-trained on ImageNet and fine-tuned on the Moments in Time dataset. We sample 16 equi-distant frames for each video, feed them to the network, and extract the \textbf{conv4} and \textbf{softmax} vectors from the CNN as the visual features of the video.

For \textbf{emb}, we encode each of the 339 category labels as a vector, more specifically a 300-dimensional CBOW word2vec embedding~\cite{MikolovEtAl:2013}. In the case of multiword phrases, we average the embeddings for each word in the phrase. For each video and for each category label, we scale the category embedding elementwise by its corresponding CNN softmax posterior prediction. 

For each video segment in our experiments, \textbf{conv4} is represented as a $7 \times 7 \times 2048$ matrix and \textbf{emb} as a $339 \times 300$ matrix. The former can be interpreted as 49 video region summaries, where each region is a cell of a $7 \times 7$ grid that divides the video spatially. The latter can be seen as a description of the video segment based only on the 339 action categories.

\section{MMT model}

We base our model on the {\bf transformer architecture}~\cite{vaswani2017attention} for neural machine translation. Our architecture is a multi-layer encoder-decoder using the \texttt{tensor2tensor}\footnote{\url{https://github.com/tensorflow/tensor2tensor}}~\cite{vaswani2018tensor2tensor} library. 

The encoder and decoder blocks are as follows: 

\paragraph{Encoder Block} ($\mathcal{E}$):  The encoder block comprises $6$ layers, with each containing two sublayers of multi-head self-attention mechanism followed by a fully connected feed forward neural network. We follow the standard implementation and employ residual connections between each layer, as well as layer normalisation. The output of the encoder forms the encoder memory which consists of contextualised representations for each of the source tokens ($M_{\mathcal{E}}$).

\paragraph{Decoder Block} ($\mathcal{D}$): The decoder block also comprises $6$ layers. It contains an additional sublayer which performs multi-head attention over the outputs of the encoder block. Specifically, decoding layer $d_{l_i}$ is the result of a) multi-head attention over the outputs of the encoder which in turn is a function of the encoder memory and the outputs from the previous layer: $A_{\mathcal{D}{\rightarrow}\mathcal{E}} = f(M_{\mathcal{E}}, d_{l_{i-1}})$ where, the keys and values are the encoder outputs and the queries correspond to the decoder input, and b) the multi-head self attention which is a function of the generated outputs from the previous layer: $A_{\mathcal{D}} = f(d_{l_{i-1}})$. 

Our multimodal transformer models follow one of the two formulations below for conditioning translations on image information:

\begin{itemize}

\item {\bf Additive image conditioning (AIC)} The 2048-D \textbf{videosum} feature vector is projected and then added to each of the outputs of the encoder. The projection matrix is jointly learned with the model.
\item {\bf Attention over image features (AIF)} The model attends over image features, as in 
\citet{HelclEtAl:2018}, where the decoder block now contains an additional cross-attention sub-layer 
$A_{\mathcal{D\rightarrow}\mathcal{V}}$ which attends to the visual information. The keys and values correspond to the visual information.

For \textbf{conv4} and \textbf{emb}, the attention is distributed across the 49 video regions and the 339 action categories, respectively. For \textbf{videosum}, the 2048-D feature vector is reshaped in row-major order into a $32 \times 64$ matrix,  so that the attention is over the 32 rows.
\end{itemize}

\paragraph{Training} 
We keep the hyperparameter settings as in \citet{ive2019distil}, i.e. we use the \texttt{transformer\_big} parameter set with 16 heads, a hidden state size of 1024, a base learning rate of 0.05, and a dropout rate of 0.1 for layer pre- and post-processing at training time. We optimise our models  with cross entropy loss and  Adam as optimiser~\cite{kingma2014adam}. Training is performed until convergence.\footnote{We use early stopping with patience of 10 epochs based on the validation \bleu score.} We optimise the number of warmup steps during the multi-GPU training according to \citet{popel2018training}. We apply beam search of size 10 and alpha of 1.0 for inference.

\begin{figure*}[!h]
  \begin{subfigure}[c]{\textwidth}
  \vspace{1em}
  \begin{small}
    \begin{tabular}{c p{1.5cm}p{10cm}}
      \multirow{3}[15]{*}{\includegraphics[width=0.20\textwidth]{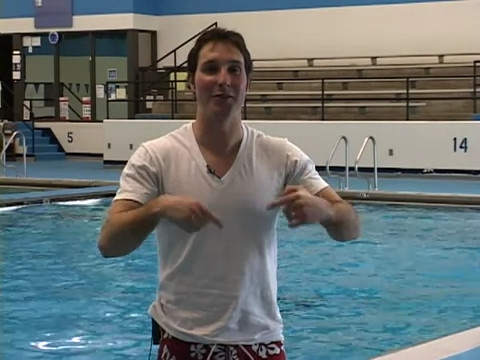}} & EN & So, how do I make sure that I \underline{spin} all the way around, or, how do you make sure? \\[1ex]
      & {\tt text-only}  & Ent\~{a}o, como eu me certifico de \underline{cortar} a toda a volta, ou como voc\^e se certifica?\\[1ex]
      & \conv & Ent\~{a}o, como eu me certifico de \underline{dar a volta}, ou, como voc\^e se certifica? \\[1ex]
      & \emb & Ent\~{a}o, como eu me certifico de que eu \underline{viro} todo o caminho, ou, como voc\^e se certifica?\\[1ex]
      & PT & Ent\~{a}o, como eu me certifico de \underline{girar} ao redor, ou, como voc\^e se certifica?\\[1ex]
  \end{tabular}
  \end{small}
  \caption{\conv guesses the masked word \underline{spin} correctly as \textit{dar a volta}, while the {\tt text-only} model translates it incorrectly as \textit{cortar} (cut) and \emb translates it partially correctly as \textit{virar} (turn)}
  \end{subfigure}
  \begin{subfigure}[c]{\textwidth}
  \vspace{1em}
    \begin{small}
    \begin{tabular}{c p{1.7cm}p{9.6cm}}
      \multirow{3}[15]{*}{\includegraphics[width=0.20\textwidth]{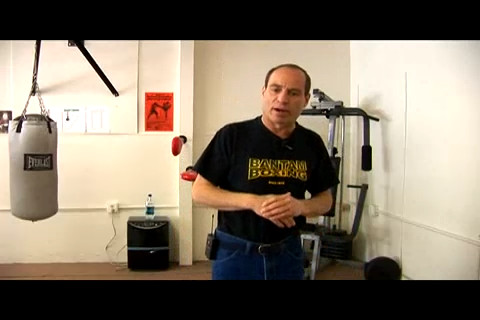}} & EN & In this clip we're \underline{talking} about footwork, we're going to be \underline{covering} the moving forward aspect of it. \\[1ex]
      & {\tt text-only}  & Neste clipe, estamos \underline{falando} de trabalho de p\'{e}s, vamos \underline{discutir} o aspecto da frente dele.\\[1ex]
      & \conv &  Neste clipe, estamos \underline{falando} sobre o trabalho de p\'{e}s, vamos \underline{cobrir} o aspecto da mudan\c{c}a para a frente. \\[1ex]
      & \emb & Neste clipe, estamos \underline{falando} de footwork, vamos \underline{discutir} o aspecto do movimento em movimento. \\[1ex]
      & PT & Neste pequeno v\'{i}deo, estamos \underline{falando} de trabalho de p\'{e}s, vamos estar \underline{cobrindo} o aspecto avan\c{c}ado.\\[1ex]
  \end{tabular}
  \end{small}
  \caption{\conv guesses \underline{talking} and \underline{covering} correctly as \textit{falar} (talk) and \textit{cobrir} (cover); the other models get the first word right, but translate the other word as  \textit{discutir} (discuss)}
  \end{subfigure}
    \begin{subfigure}[c]{\textwidth}
  \vspace{1em}
    \begin{small}
    \begin{tabular}{c p{1.7cm}p{9.6cm}}
      \multirow{3}[15]{*}{\includegraphics[width=0.20\textwidth]{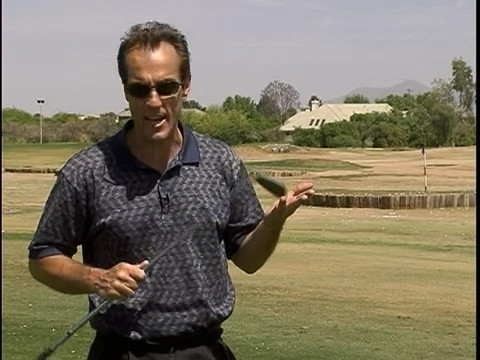}} & EN & I might use the sixty degree wedge a bit too, but the sand wedge obviously is useful for getting out of the ruff, \underline{hitting} the ball from the fairway, getting out of sand. \\[1ex]
      & {\tt text-only} & Eu poderia usar a cunha de sessenta graus tamb\'{e}m, mas a cunha de areia, obviamente, \'{e} \'{u}til para sair do ruff, \underline{tirar} a bola do fairway, sair da areia.\\[1ex]
      & \conv &  Eu poderia usar a cunha de sessenta graus um pouco tamb\'{e}m, mas a cunha de areia obviamente \'{e} \'{u}til para sair do pesco\c{c}o,  \underline{bater} na bola do fairway, saindo da areia.\\[1ex]
      & \emb & Eu posso usar a cunha de sessenta graus um pouco tamb\'{e}m, mas a cunha de areia obviamente \'{e} \'{u}til para sair do ruff,  \underline{bater} a bola do fairway, sair da areia.\\[1ex]
      & PT & Eu tamb\'{e}m poderia usar a wedge de sessenta graus, mas a sand-wedge, obviamente, \'{e} útil para sair do ruff, \underline{acertar} a bola no fairway, sair da areia.\\[1ex]
  \end{tabular}
  \end{small}
  \caption{\conv and \emb guess \underline{hit} correctly as \textit{bater} (the gold-standard \textit{acertar} has the same meaning this example), while the \textit{text-only} model  translates it as \textit{tirar} (remove)}
  \end{subfigure}

\caption{\label{fig:main_ex} Examples of improvements of \conv and \emb over the text-only baseline. Underlined text denotes masked words and their translations.}
\end{figure*}

\section{Results}

Table \ref{with_punct_results_table} reports the results\footnote{The scores may not be comparable to the ones on the How2 Machine Translation Challenge leaderboard, since we do not remove punctuation from the model inferences, but it is the case with submissions to the challenge. The same caveat applies to Table \ref{with_punct_incongruent_results_table}.} of our experiments using \bleu \cite{Papieni02bleu} as metric.\footnote{We measure the performance  with Multeval \cite{clark-etal-2011-better}. We use tokenised and lowercased reference and hypotheses, as in the official challenge.}

\begin{table}[t]
\caption{Results for the test set. We report \bleu scores. Bold highlights our best results.}
\label{with_punct_results_table}
\vskip 0.15in
\begin{center}
\begin{small}
\begin{sc}
\begin{tabular}{lcccr}
\toprule
Setup           & \org   & \act  & \all \\
\midrule
{\tt text-only}          & 55.9          & 53.6          & 44.1    \\
\aic               & 55.6           & 53.6          & 44.2    \\
\videosum      & 55.7           & 53.3          & 44.0    \\
\conv          & 55.6           & \bf 53.8          & 44.4    \\
\emb      & \bf 56.2           & 53.5          & \bf 44.5    \\
\bottomrule
\end{tabular}
\end{sc}
\end{small}
\end{center}
\vskip -0.1in
\end{table}

\begin{table}[t]
\caption{\bleu deltas recorded when models are given incongruent visual features, compared to the scores reported in Table \ref{with_punct_results_table}}
\label{with_punct_incongruent_results_table}
\vskip 0.15in
\begin{center}
\begin{small}
\begin{sc}
\begin{tabular}{lcccr}
\toprule
Setup           & \org   & \act  & \all \\
\midrule
\aic               & $\uparrow 0.1$          & $\downarrow 0.7$         & $\downarrow 1.0$    \\
\videosum      & $\downarrow 0.1$          & $\downarrow 0.3$      & $\downarrow 0.5$   \\
\conv          & $\downarrow 0.3$         & $\downarrow 0.5$         & $\downarrow 0.8$   \\
\emb      & $\downarrow 0.1$       & $\downarrow 0.4$         & $\downarrow 0.3$   \\
\bottomrule
\end{tabular}
\end{sc}
\end{small}
\end{center}
\vskip -0.1in
\end{table}

Our unmasked {\sc Text-Only} baseline achieves a \bleu score of 55.9. As expected, the scores for masked models are lower: for \act, we observe a \bleu of 53.6; for \all a \bleu of 44.1.

Overall, \aic is on par with {\sc Text-Only} for \act and \all. For \org, a drop of 0.3 \bleu is recorded.

\videosum, on the other hand, leads to degraded performance: a 0.2 decrease for \org, 0.3 for \act, and 0.1 for \all. This reveals the possibility that a single global feature vector is better used as a whole as opposed to as segmented sub-vectors for attention.

\textbf{conv4} and \textbf{emb} features contribute to our best-performing models for different masking settings. \conv achieves deltas of -0.3, 0.2 and 0.3 points over the text-only model respectively for \org, \act and \all. The figures for \emb models are 0.3, -0.1 and 0.4 \bleu points on the same settings. The above also means that \conv is our best for \act and \emb takes the first place for \org and \all.

Overall, the top-performing models achieve modest improvements over the text-only baseline in each masking setting. To probe into how much visual features matter to the multimodal models, we conduct incongruent decoding \citet{naacl19-probing}, where we feed the visual features in reverse order. The assumption is that models that have learned to exploit visual information to aid translation will suffer from the mismatch and hence result in performance deterioration. We summarise the results of our incongruent decoding for the multimodal models in Table \ref{with_punct_incongruent_results_table}.

As expected, visual incongruence leads to worse results (as much as a 1.0 \bleu drop) in almost all the settings, which proves that multimodality indeed exerts positive influence on the translation. Also evident from the table is that the score changes are considerably more pronounced for \act and \all than for \org, pointing to the possibility that the verb masking leads to the visual features being more relied upon for translation and that a mask-free source sentence may be sufficient already for quality translation.

To sum up, a major finding is the importance of the visual features to the multimodal translation despite the generally moderate improvements achieved in terms of the automatic metric. Surprisingly, the delta in \bleu between text-only and multimodal models for the masked datasets is not substantially larger than for the original dataset (0.3 \bleu points). It is smaller for \act (0.2 \bleu points), and similar for \all (0.4 \bleu points).

Additionally, we note that in none of the model settings are the video features able to help bridge the gap between \org and \act performances, not even in \conv and \emb where the visual features are from a CNN fine-tuned for classifying videos into classes whose labels are closely related to the verbs masked in \act. However, the gap between \org and \all is slightly smaller for some multimodal models than for the text-only model.  

\subsection{Human Analysis}

Automatic metrics often fail to capture nuances in translation quality, such as the ones we expect the visual modality to help with, which -- according to human perception -- lead to better translations \cite{elliott-EtAl:2017:WMT, BarraultEtAl:2018}.  We thus performed human evaluation of our best outputs involving native speakers of Portuguese (four annotators) who are fluent speakers of English. We focused only on the evaluation for \act.

The annotators were asked to rank randomly selected test samples according to how well they convey the meaning of the source (50 samples per annotator). For each source segment, the annotator was shown the outputs of three systems: {\tt text-only}, \conv and \emb. They also had access to reference translations. A rank could be assigned from 1 to 3, allowing ties~\cite{W17-4717}. Annotators could assign zero rank to all translations if they were judged incomprehensible.

Following the common practice in human evaluation for many machine translation shared tasks \cite{W17-4717}, each system was then assigned a score which reflects the proportion of times it was judged to be better than or equal to the other two systems.

Table~\ref{table:human_res} shows the human evaluation results. Contrary to the automatic evaluation results for \act, the \emb setup is generally favoured by human preference. {\tt text-only} and \conv demonstrate similar performance.

Figure~\ref{fig:main_ex} illustrates some cases where \emb or \conv outperforms the text-only model.

\begin{table}[!h]
\begin{center}
\scalebox{0.73}{
\begin{tabular}{c c c}
\toprule
{\tt text-only} & \conv & \emb \\ \midrule
0.75 & 0.73 & 0.81 \\
\bottomrule
\end{tabular}}
\end{center}
\caption{\label{table:human_res} Human ranking results for \act: micro-averaged rank over four annotators.}
\end{table}

\section{Conclusions}

We investigated a state of the art multimodal machine translation approach on the How2 dataset. Our focus was on exploring visual features that attempt to represent action information, and on probing their contribution when the input text is corrupted to remove action-related words. The hypothesis was that a well designed multimodal model based on informative visual features should be able to recover from the lack of textual information by leveraging the visual information. Our main results are as follows: (i) performance improvements over the text-only baseline can be achieved by the multimodal models, and the best in all three masking settings are produced by the models that exploit visual features from an action classification network; (ii) visual information is important to the multimodal models, especially in the verb-masked cases where substantial performance drops are observed in case of a visual feature mismatch; (iii) human evaluation indicates that representing features of verb-related actions with word embeddings to exploit similarities between respective verbs could be beneficial. 
These are promising results for multimodal machine translation and for the use of action-related visual features in this context.

\section*{Acknowledgement}

This work was supported by the MultiMT (H2020 ERC Starting Grant No. 678017) and MMVC (Newton Fund Institutional Links Grant, ID 352343575)
projects. We also thank the annotators for their valuable help.

\nocite{langley00}

\bibliography{how2}

\begin{thebibliography}{20}
\providecommand{\natexlab}[1]{#1}
\providecommand{\url}[1]{\texttt{#1}}
\expandafter\ifx\csname urlstyle\endcsname\relax
  \providecommand{\doi}[1]{doi: #1}\else
  \providecommand{\doi}{doi: \begingroup \urlstyle{rm}\Url}\fi

\bibitem[Barrault et~al.(2018)Barrault, Bougares, Specia, Lala, Elliott, and
  Frank]{BarraultEtAl:2018}
Barrault, L., Bougares, F., Specia, L., Lala, C., Elliott, D., and Frank, S.
\newblock Findings of the third shared task on multimodal machine translation.
\newblock In \emph{Proceedings of the {{Third Conference}} on {{Machine
  Translation}}: {{Shared Task Papers}}}, pp.\  304--323. {Association for
  Computational Linguistics}, 2018.
\newblock URL \url{http://aclweb.org/anthology/W18-6402}.

\bibitem[Bojar et~al.(2017)Bojar, Chatterjee, Federmann, Graham, Haddow, Huang,
  Huck, Koehn, Liu, Logacheva, Monz, Negri, Post, Rubino, Specia, and
  Turchi]{W17-4717}
Bojar, O., Chatterjee, R., Federmann, C., Graham, Y., Haddow, B., Huang, S.,
  Huck, M., Koehn, P., Liu, Q., Logacheva, V., Monz, C., Negri, M., Post, M.,
  Rubino, R., Specia, L., and Turchi, M.
\newblock Findings of the 2017 conference on machine translation (wmt17).
\newblock In \emph{Proceedings of the Second Conference on Machine
  Translation}, pp.\  169--214. Association for Computational Linguistics,
  2017.
\newblock \doi{10.18653/v1/W17-4717}.
\newblock URL \url{http://aclweb.org/anthology/W17-4717}.

\bibitem[Caglayan et~al.(2019)Caglayan, Madhyastha, Specia, and
  Barrault]{naacl19-probing}
Caglayan, O., Madhyastha, P., Specia, L., and Barrault, L.
\newblock Probing the need for visual context in multimodal machine
  translation.
\newblock \emph{CoRR}, abs/1903.08678, 2019.
\newblock URL \url{http://arxiv.org/abs/1903.08678}.

\bibitem[Clark et~al.(2011)Clark, Dyer, Lavie, and
  Smith]{clark-etal-2011-better}
Clark, J.~H., Dyer, C., Lavie, A., and Smith, N.~A.
\newblock Better hypothesis testing for statistical machine translation:
  Controlling for optimizer instability.
\newblock In \emph{Proceedings of the 49th Annual Meeting of the Association
  for Computational Linguistics: Human Language Technologies}, pp.\  176--181,
  Portland, Oregon, USA, June 2011. Association for Computational Linguistics.
\newblock URL \url{https://www.aclweb.org/anthology/P11-2031}.

\bibitem[Elliott et~al.(2016)Elliott, Frank, Sima'an, and
  Specia]{elliott-etall_VL:2016}
Elliott, D., Frank, S., Sima'an, K., and Specia, L.
\newblock Multi30k: Multilingual english-german image descriptions.
\newblock In \emph{5th Workshop on Vision and Language}, pp.\  70--74, Berlin,
  Germany, 2016.
\newblock URL \url{http://aclweb.org/anthology/W16-3210}.

\bibitem[Elliott et~al.(2017)Elliott, Frank, Barrault, Bougares, and
  Specia]{elliott-EtAl:2017:WMT}
Elliott, D., Frank, S., Barrault, L., Bougares, F., and Specia, L.
\newblock Findings of the second shared task on multimodal machine translation
  and multilingual image description.
\newblock In \emph{Proceedings of the Second Conference on Machine Translation,
  Volume 2: Shared Task Papers}, pp.\  215--233, Copenhagen, Denmark, September
  2017. Association for Computational Linguistics.
\newblock URL \url{http://www.aclweb.org/anthology/W17-4718}.

\bibitem[Hara et~al.(2018)Hara, Kataoka, and Satoh]{hara2018can}
Hara, K., Kataoka, H., and Satoh, Y.
\newblock Can spatiotemporal 3d cnns retrace the history of 2d cnns and
  imagenet.
\newblock In \emph{Proceedings of the IEEE conference on Computer Vision and
  Pattern Recognition}, pp.\  6546--6555, 2018.

\bibitem[Helcl et~al.(2018)Helcl, Libovick\'y, and Varis]{HelclEtAl:2018}
Helcl, J., Libovick\'y, J., and Varis, D.
\newblock {{CUNI}} system for the {{WMT18}} multimodal translation task.
\newblock In \emph{Proceedings of the {{Third Conference}} on {{Machine
  Translation}}: {{Shared Task Papers}}}, pp.\  616--623. {Association for
  Computational Linguistics}, 2018.
\newblock URL \url{http://aclweb.org/anthology/W18-6441}.

\bibitem[Ive et~al.(2019)Ive, Madhyastha, Swaroop, and Specia]{ive2019distil}
Ive, J., Madhyastha, Swaroop, P., and Specia, L.
\newblock {Distilling Translations with Visual Awareness}.
\newblock In \emph{{Proceedings of the 57th Annual Meeting of the Association
  for Computational Linguistics}}, 2019.

\bibitem[Kingma \& Ba(2014)Kingma and Ba]{kingma2014adam}
Kingma, D.~P. and Ba, J.
\newblock Adam: A method for stochastic optimization.
\newblock \emph{arXiv preprint arXiv:1412.6980}, 2014.

\bibitem[Mikolov et~al.(2013)Mikolov, Chen, Corrado, and
  Dean]{MikolovEtAl:2013}
Mikolov, T., Chen, K., Corrado, G., and Dean, J.
\newblock Efficient estimation of word representations in vector space.
\newblock In Bengio, Y. and LeCun, Y. (eds.), \emph{Proceedings of the 1st
  {{International Conference}} on {{Learning Representations}}, {{ICLR}} 2013,
  {{Workshop Track Proceedings}}}, Scottsdale, AZ, USA, May 2013.
\newblock URL \url{http://arxiv.org/abs/1301.3781}.

\bibitem[Monfort et~al.(2019)Monfort, Andonian, Zhou, Ramakrishnan, Bargal,
  Yan, Brown, Fan, Gutfruend, Vondrick, et~al.]{monfortmoments}
Monfort, M., Andonian, A., Zhou, B., Ramakrishnan, K., Bargal, S.~A., Yan, T.,
  Brown, L., Fan, Q., Gutfruend, D., Vondrick, C., et~al.
\newblock Moments in time dataset: one million videos for event understanding.
\newblock \emph{IEEE Transactions on Pattern Analysis and Machine
  Intelligence}, pp.\  1--8, 2019.
\newblock ISSN 0162-8828.
\newblock \doi{10.1109/TPAMI.2019.2901464}.

\bibitem[Papineni et~al.(2002)Papineni, Roukos, Ward, and Zhu]{Papieni02bleu}
Papineni, K., Roukos, S., Ward, T., and Zhu, W.-J.
\newblock Bleu: a method for automatic evaluation of machine translation.
\newblock In \emph{Proceedings of 40th Annual Meeting of the Association for
  Computational Linguistics}, pp.\  311--318, 2002.
\newblock URL \url{http://www.aclweb.org/anthology/P02-1040}.

\bibitem[Popel \& Bojar(2018)Popel and Bojar]{popel2018training}
Popel, M. and Bojar, O.
\newblock Training tips for the transformer model.
\newblock \emph{The Prague Bulletin of Mathematical Linguistics}, 110\penalty0
  (1):\penalty0 43--70, 2018.

\bibitem[Sanabria et~al.(2018)Sanabria, Caglayan, Palaskar, Elliott, Barrault,
  Specia, and Metze]{how2-dataset}
Sanabria, R., Caglayan, O., Palaskar, S., Elliott, D., Barrault, L., Specia,
  L., and Metze, F.
\newblock How2: {A} large-scale dataset for multimodal language understanding.
\newblock \emph{CoRR}, abs/1811.00347, 2018.
\newblock URL \url{http://arxiv.org/abs/1811.00347}.

\bibitem[Sennrich et~al.(2015)Sennrich, Haddow, and Birch]{sennrich2015neural}
Sennrich, R., Haddow, B., and Birch, A.
\newblock Neural machine translation of rare words with subword units.
\newblock \emph{arXiv preprint arXiv:1508.07909}, 2015.

\bibitem[Specia et~al.(2016)Specia, Frank, Sima'an, and
  Elliott]{SpeciaEtAl:2016}
Specia, L., Frank, S., Sima'an, K., and Elliott, D.
\newblock A shared task on multimodal machine translation and crosslingual
  image description.
\newblock In \emph{Proceedings of the First Conference on Machine Translation},
  pp.\  543--553. Association for Computational Linguistics, 2016.
\newblock \doi{10.18653/v1/W16-2346}.
\newblock URL \url{http://www.aclweb.org/anthology/W16-2346}.

\bibitem[Vaswani et~al.(2017)Vaswani, Shazeer, Parmar, Uszkoreit, Jones, Gomez,
  Kaiser, and Polosukhin]{vaswani2017attention}
Vaswani, A., Shazeer, N., Parmar, N., Uszkoreit, J., Jones, L., Gomez, A.~N.,
  Kaiser, {\L}., and Polosukhin, I.
\newblock Attention is all you need.
\newblock In \emph{Advances in Neural Information Processing Systems}, pp.\
  5998--6008, 2017.

\bibitem[Vaswani et~al.(2018)Vaswani, Bengio, Brevdo, Chollet, Gomez, Gouws,
  Jones, Kaiser, Kalchbrenner, Parmar, et~al.]{vaswani2018tensor2tensor}
Vaswani, A., Bengio, S., Brevdo, E., Chollet, F., Gomez, A.~N., Gouws, S.,
  Jones, L., Kaiser, {\L}., Kalchbrenner, N., Parmar, N., et~al.
\newblock Tensor2tensor for neural machine translation.
\newblock \emph{arXiv preprint arXiv:1803.07416}, 2018.

\bibitem[Xie et~al.(2017)Xie, Girshick, Doll{\'a}r, Tu, and
  He]{xie2017aggregated}
Xie, S., Girshick, R., Doll{\'a}r, P., Tu, Z., and He, K.
\newblock Aggregated residual transformations for deep neural networks.
\newblock In \emph{Proceedings of the IEEE conference on computer vision and
  pattern recognition}, pp.\  1492--1500, 2017.

\end{thebibliography}
\bibliographystyle{icml2019}

\end{document}